\documentclass[runningheads]{llncs}
\usepackage{graphicx}
\usepackage{comment}
\usepackage{amsmath,amssymb} %
\usepackage{color}

\usepackage{hyperref}
\usepackage{url}
\usepackage{algorithm}
\usepackage{algorithmicx}
\usepackage{algpseudocode}
\usepackage{multirow}
\usepackage{graphicx}
\usepackage{booktabs}
\usepackage{cite}

\usepackage{placeins}
\usepackage{caption}
\usepackage{subcaption}

\def\onedot{.}
\def\eg{\emph{e.g}\onedot} 
\def\ie{\emph{i.e}\onedot}

\def\wrt{w.r.t\onedot}

\begin{document}
\pagestyle{headings}
\mainmatter

\title{BigNAS: Scaling Up Neural Architecture Search with Big Single-Stage Models}

\titlerunning{BigNAS: Neural Architecture Search with Big Single-Stage Models}

\author{Jiahui Yu\inst{1,2} \and
Pengchong Jin\inst{1} \and
Hanxiao Liu\inst{1} \and
Gabriel Bender\inst{1} \and\\
Pieter-Jan Kindermans\inst{1} \and
Mingxing Tan\inst{1} \and
Thomas Huang\inst{2} \and
Xiaodan Song\inst{1} \and
Ruoming Pang\inst{1} \and
Quoc Le\inst{1}
}

\authorrunning{J. Yu et al.}

\institute{Google Brain \and University of Illinois at Urbana-Champaign \\\email{jiahuiyu@google.com}}

\maketitle

\begin{abstract}
Neural architecture search (NAS) has shown promising results discovering models that are both accurate and fast. For NAS, training a \textit{one-shot model} has become a popular strategy to rank the relative quality of different architectures (\textit{child models}) using a single set of shared weights. However, while one-shot model weights can effectively \textit{rank} different network architectures, the absolute accuracies from these shared weights are typically far below those obtained from stand-alone training. To compensate, existing methods assume that the weights must be retrained, finetuned, or otherwise post-processed after the search is completed. These steps significantly increase the compute requirements and complexity of the architecture search and model deployment. In this work, we propose BigNAS, an approach that challenges the conventional wisdom that post-processing of the weights is necessary to get good prediction accuracies. \emph{Without extra retraining or post-processing steps}, we are able to train a single set of shared weights on ImageNet and use these weights to obtain child models whose sizes range from 200 to 1000 MFLOPs. Our discovered model family, BigNASModels, achieve top-1 accuracies ranging from 76.5\% to 80.9\%, surpassing state-of-the-art models in this range including EfficientNets and Once-for-All networks without extra retraining or post-processing. We present ablative study and analysis to further understand the proposed BigNASModels.
\end{abstract}

\section{Introduction}
Designing network architectures that are both accurate and efficient is crucial for deep learning on edge devices. A single neural network architecture can require more than an order of magnitude more inference time if it is deployed on a slower device \cite{yu2018slimmable}. Furthermore, even two devices which have similar overall speeds (\eg, phone CPUs made by different manufacturers) can favor very different network architectures due to hardware and device driver differences \cite{wu2019fbnet}. This makes it appealing to not only search for architectures of varying sizes that are optimized for specific devices, but also ensure that these models can be deployed effectively.

In the past, to optimize network architectures for a single device and latency target~\cite{tan2018mnasnet}, Neural Architecture Search (NAS) methods \cite{zoph2016neural,zoph2018learning,real2018regularized} have shown to be effective. While early NAS methods were prohibitively expensive for most practitioners, recent \emph{efficient} NAS methods based on weight sharing  reduce search costs by orders of magnitude \cite{pham2018efficient,bender2018understanding,liu2018darts,yu2019network}. These methods work by training a \emph{super-network} and then identifying a path through the network -- a subset of its operations -- which gives the best possible accuracy while satisfying a user-specified latency constraint for a specific hardware device. The advantage of this approach is that we can train the super-network and then use it to \textit{rank} many different candidate architectures from a user-defined search space.

However, the \textit{absolute accuracies} of predictions obtained from this super-network are typically much lower than those of models trained from scratch in stand-alone fashion~\cite{bender2018understanding}. For this reason, it is commonly assumed that significant post-processing of the super-network's weights is necessary to obtain high-quality accuracies for model deployment. For example, one proposed solution is to retrain a separate model for each device of interest and each latency budget of interest~\cite{wu2019fbnet, cai2018proxylessnas}. However, this incurs significant overhead, especially if the number of deployment scenarios is large. A second solution would be to post-process the weights after training is finished; for example, using the \textit{progressive shrinking} heuristic proposed for Once-for-All networks \cite{cai2019once}. However, this post-processing step complicates the model training pipeline. Moreover, the child models from Once-for-All networks \cite{cai2019once} still requires fine-tuning with additional epochs (\eg, 75 epochs on ImageNet) to achieve the best accuracies.

In this work, we reassess the popular belief that the retraining or post-processing of the shared weights is necessary in order to obtain competitive accuracies. We propose several techniques to bridge the gap between the distinct initialization and learning dynamics across small and big child models with shared parameters. With these techniques, we are able to train a \textit{single-stage model}: a single model from which we can directly slice high-quality child models \emph{without any extra post-processing}.

\begin{figure}[tb!]
\centering
\includegraphics[width=\linewidth]{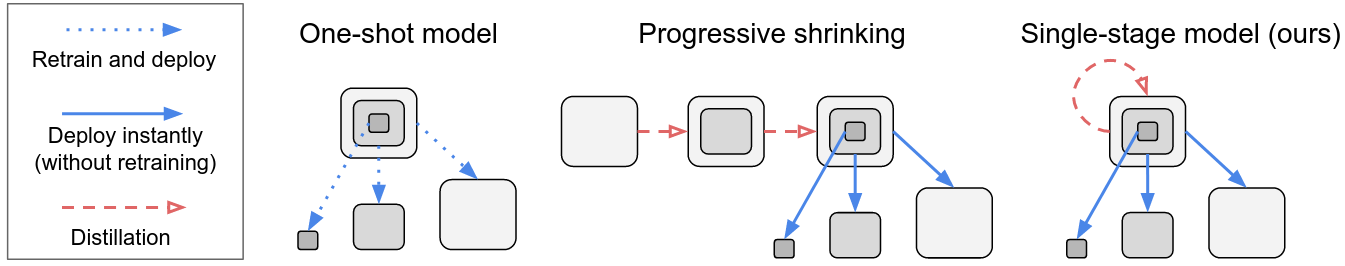}
\caption{Comparison with several existing workflows. We use nested squares to denote models with shared weights, and use the size of the square to denote the size of each model. Workflow in the middle refers the concurrent work from \cite{cai2019once}, where submodels are sequentially induced through progressive distillation and channel sorting. We simultaneously train all child models in a single-stage model with proposed modifications, and deploy them without retraining or finetuning.
}
\label{figs:schematic}
\end{figure}

We search over a big \emph{single-stage} model that contains both small child models ($\sim$200 MFLOPs, comparable to MobileNetV3) and big child models ($\sim$1 GFLOPs, comparable to EfficientNets). Different from existing one-shot methods~\cite{bender2018understanding,liu2018darts,brock2018smash,pham2018efficient,yu2019network}, our trained single-stage model offers a much wider coverage of model capacities, and more importantly, all child models are trained in a way such that they simultaneously reach excellent performance at the end of the search phase. Architecture selection can be then carried out via a simple coarse-to-fine selection strategy. Once an architecture is selected, we can obtain a child model by simply slicing the single-stage model for instant deployment w.r.t.\ the given constraints such as memory footprint and/or runtime latency. The workflow is illustrated in Figure~\ref{figs:schematic}.

The success of simplified BigNAS workflow relies on a single objective: how to train a high-quality single-stage model? This objective is challenging on its own. For example, we find that the training loss explodes if a big single-stage model is not properly initialized; during the training process, big child models start to overfit before small ones plateau; empirically, bigger child models tend to overfit more on the training data. To address these challenges, we systematically study and revisit conventional training techniques of stand-alone networks, and adapt them to train weight-sharing single-stage models. With the proposed techniques, we are able train a high-quality single-stage model on ImageNet and obtain a family of child models that simultaneously surpass all the state-of-the-art models in the range of 200 to 1000 MFLOPs, including EfficientNets B0-B2 (1.6\% more accurate under 400 MFLOPs), without retraining or finetuning the child models upon the completion of search. For example, one of our child models achieves 80.9\% top-1 accuracy at 1G FLOPs (\(4 \times\) less computation than a ResNet-50).

\section{Related Work}
Earlier NAS methods~\cite{zoph2016neural, zoph2018learning, liu2017hierarchical, liu2018progressive, real2018regularized} train thousands of candidate architectures from scratch (on a smaller proxy task) and use their validation performance as the feedback to an algorithm that learns to focus on the most promising regions in the search space. More recent works have sought to amortize the cost by training a single over-parameterized \emph{one-shot model}. Each architecture in the search space uses only a subset of the operations in the one-shot model; these \emph{child models} can be efficiently ranked by using the shared weights to estimate their relative accuracies~\cite{brock2018smash, pham2018efficient, bender2018understanding, liu2018darts, cai2018proxylessnas, wu2019fbnet, yu2019network, hu2020dsnas, Yang_2020_CVPR}.

As a complementary direction, resource-aware NAS methods are proposed to simultaneously maximize prediction accuracy and minimize resource requirements such as latency, FLOPs, or memory footprints~\cite{tan2018mnasnet, cai2019once, wu2019fbnet, stamoulis2019single, guo2019single, yu2019network, Berman_2020_CVPR}.

All the aforementioned approaches require two-stage training: once the best architectures have been identified (either through the proxy tasks or using a one-shot model), they have to be retrained from scratch to obtain a final model with higher accuracy. In most of these existing works, a single search experiment only targets a single resource budget or a narrow range of resource budgets at a time.

To alleviate these issues, \cite{cai2019once} proposed a progressive training approach (OFA) concurrently with our work. The idea is to pre-train a single full network and then progressively distill it to obtain the smaller networks. Moreover, a channel sorting procedure is required to progressively construct the smaller networks. In our proposed BigNAS, however, all the child models in the single-stage model are trained \emph{simultaneously}, allowing the learning of small and big networks to mutually benefit each other. During the training, we always keep lower-index channels in each layer and lower-index layers in each stage for our child models, eliminating the sorting procedure. Our BigNAS is able to handle a wider set of models (from 200 MFLOPs to 1 GFLOPs) and offers a better coverage over diverse deployment scenarios and varied resource budgets.

Our work shares high-level similarities with \emph{slimmable networks}~\cite{yu2018slimmable, yu2019universally, yu2019network} in terms of training a single shared set of weights which can be used for many child models. However, while slimmable networks are specialized to vary the number of channels only, we are able to handle a much larger space where many architectural dimensions (kernel and channel sizes, network depths, input resolutions) are searched simultaneously, subsuming and outperforming the manually-designed scaling heuristics in EfficientNets \cite{tan2019efficientnet}.

\section{Architecture Search with Single-Stage Models}
Our proposed method consists of two steps:
\begin{enumerate}
    \item We train a big \textit{single-stage model} from which we can directly sample or slice different architectures as \textit{child models} for instant inference and deployment. In contrast to previous works, our training is single-stage. In other words: the trained model weights from a search can be directly used for deployment, without any need to retrain them from scratch (\eg \cite{brock2018smash, pham2018efficient, bender2018understanding, liu2018darts, stamoulis2019single, guo2019single, yu2019network}) or otherwise post-process them (\eg, \cite{cai2019once}).
    \item We select architectures using a simple \textit{coarse-to-fine} selection method to find the most accurate model under the given resource constraints (\eg, FLOPs, memory footprint and/or runtime latency budgets on different devices).
\end{enumerate}

In the following, we will first systematically study how to train a \emph{high-quality single-stage model} from five aspects: network sampling during training, inplace distillation, network initialization,  convergence behavior and regularization. Then we will present a coarse-to-fine approach for efficient resource-aware architecture selection.

\subsection{Training a High-Quality Single-Stage Model}\label{secs:train_one_stage}
Training a high-quality single-stage model is important and highly non-trivial due to the distinct initialization and learning dynamics of small and big child models. In this section, we first generalize two techniques originally introduced by~\cite{yu2019universally} to simultaneously train a set of high-quality networks with different channel numbers, and show that both can be extended to handle a much larger space where the architectural dimensions, including kernel sizes, channel numbers, input resolutions, network depths are jointly searched. We then present three additional techniques to address the distinct initialization and learning dynamics of small and big child models.

\textbf{Sandwich Rule.} In each training step, given a mini-batch of data, the sandwich rule~\cite{yu2019universally} samples the smallest child model, the biggest (full) child model and $N$ randomly sampled child models ($N=2$ in our experiments). It then aggregates the gradients from all sampled child models before updating the weights of the single-stage model. As multiple architectural dimensions are included in our search space, the ``smallest'' child model is the one with lowest input resolution, thinnest width, shallowest depth, and smallest kernel size (the kernel of the depthwise convolutions in each inverted residual block~\cite{sandler2018inverted}). The motivation is to improve all child models in our search space simultaneously, by pushing up both the performance lower bound (the smallest child model) and the performance upper bound (the biggest child model) across all child models.

\textbf{Inplace Distillation.} During the training of a single-stage model, inplace distillation~\cite{yu2019universally} takes the soft labels predicted by the biggest possible child model (full model) to supervise all other child models. The benefit of inplace distillation comes for free in our training setting, as we always have access to the predictions of the largest child model in each gradient update step thanks to the sandwich rule. We note that all child models are only trained with the inplace distillation loss, starting from the first training step to the end of the training. The temperature hyper-parameter or the mixture of distillation/target loss~\cite{hinton2015distilling} are not used in our experiments for the sake of simplicity.

During training, input images are randomly cropped as a preliminary data augmentation step. When distilling a high-resolution teacher model into a low-resolution student model, we find that it is helpful to feed the same image patches into both the teacher and the student. In our data preparation, we first randomly crop an image with a fixed resolution (on ImageNet we use 224), and then apply bicubic interpolation to the \emph{same patch} to transform it into all target resolutions (\eg, 192, 288, 320, etc.). In this case, soft labels predicted by the biggest child model (the teacher) are more compatible with the inputs seen by other child models (the students). Therefore this can serve as a more accurate distillation signal. Our preliminary results show that this leads to \(\sim0.3\%\) improvement on average top-1 accuracy for child models compared with sampling different patches.

\textbf{Initialization.}
When we first tried to train bigger and deeper single-stage models, we found that training was highly unstable, and that the training loss exploded when we used learning rates optimized for training a normal neural network. The training started to work when we reduced the learning rate to 30\% of its original value, but this configuration lead to much worse results (\(\sim 1.0\%\) top-1 accuracy drop on ImageNet).

While stabilize model training is in general a complex and open-ended problem, we found that in this case a simple change to our setup was sufficient to stabilize training. As all child models in our search space are residual networks, we initialize the output of each residual block (before skip connection) to an all-zeros tensor by setting the learnable scaling coefficient \(\gamma = 0\) in the last Batch Normalization~\cite{ioffe2015batch} layer of each residual block, ensuring identical variance before and after each residual block regardless of the fan-in. This initialization is originally mentioned in ~\cite{goyal2017accurate} and improves accuracy by \(\sim 0.2\%\) in their setting, but is more critical in our setting (improving by \(\sim 1.0\%\)). We also add a skip connection in each stage transition when either resolutions or channels differ (using \(2 \times 2\) average pooling and/or \(1 \times 1\) convolution if necessary) to explicitly construct an identity mapping~\cite{he2016identity}.

\begin{figure}[ht]
    \centering
    \begin{subfigure}[b]{0.48\textwidth}
        \centering
        \includegraphics[width=\textwidth]{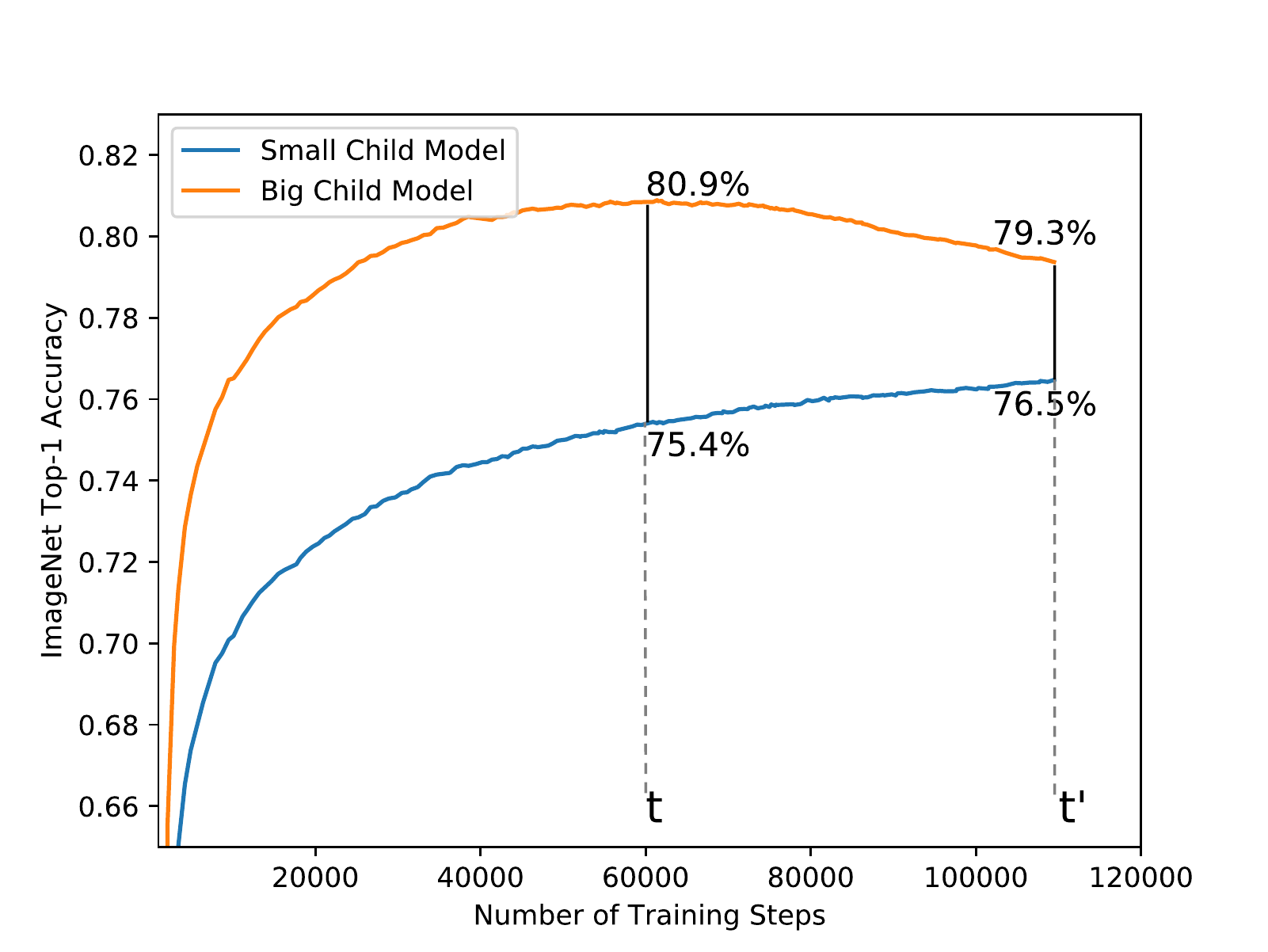}
        \caption{}
        \label{figs:convergence_a}
    \end{subfigure}
    \begin{subfigure}[b]{0.48\textwidth}
        \includegraphics[width=\textwidth]{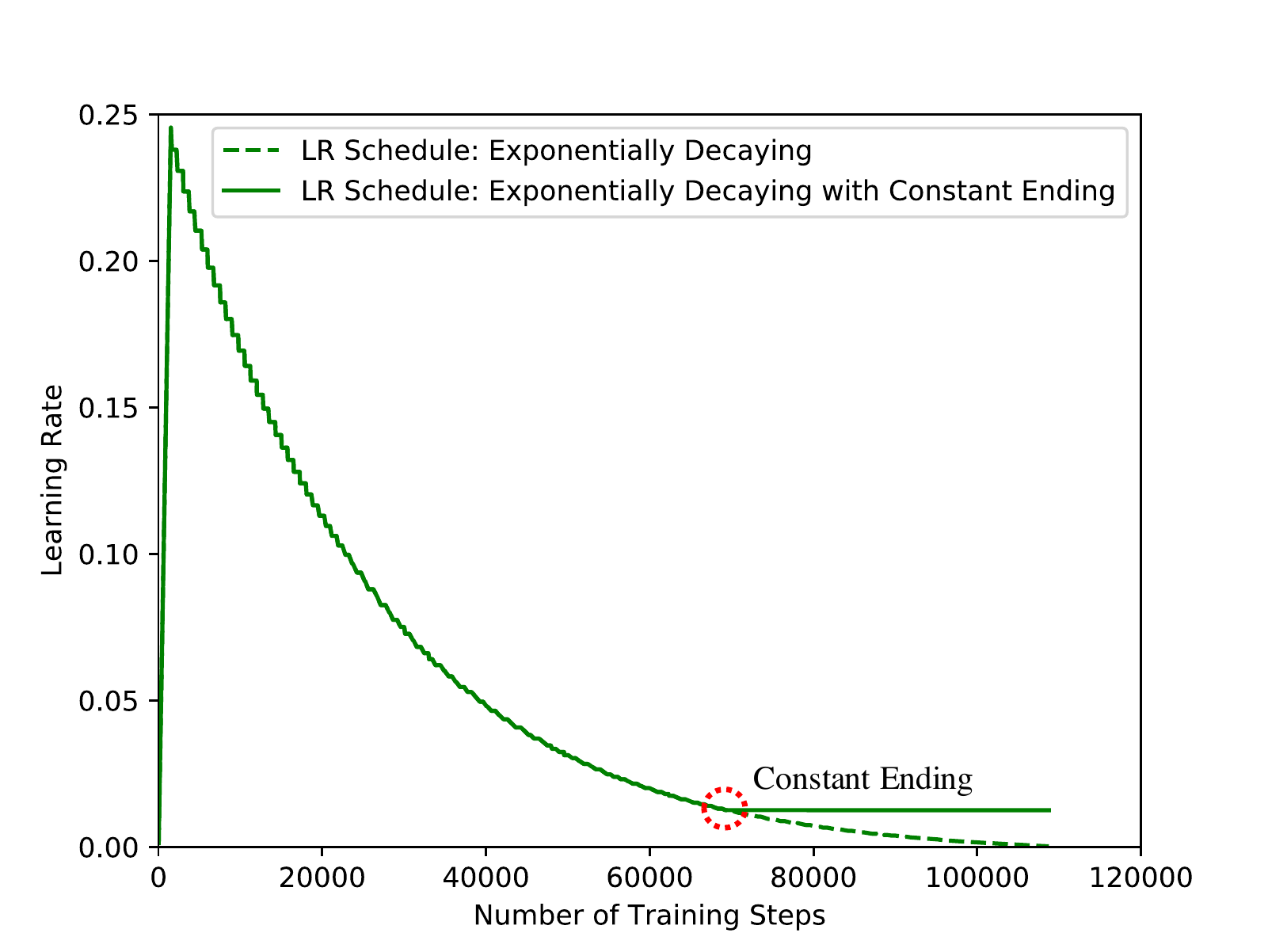}
        \caption{}
        \label{figs:convergence_b}
    \end{subfigure}
\caption{On the left, we show typical accuracy curves during the training process for both small and big child models. It reveals a common dilemma in training big single-stage models: at training step \(t\) when the performance of big child models peaks, the small child models are not fully-trained; and at training step \(t'\) when the small child models have better performance, the big child models already overfitted. On the right, we plot the simple modified learning rate schedules with constant ending to address this issue.
}
\end{figure}

\textbf{Convergence Behavior.} In practice, we find that big child models converge faster while small child models converge slower. Figure~\ref{figs:convergence_a} shows the typical learning curves during the training of a single-stage model, where we plot the validation accuracies of a small and a big child model over time. This reveals a dilemma: at training step \(t\) when the performance of big child models peaks, the small child models are not fully-trained; and at training step \(t'\) when the small child models have better performance, the big child models already overfitted.

To address this issue, we put our focus on the learning rate schedule. We first plot the optimized and widely used exponentially decaying learning rate schedule for MobileNet-series~\cite{howard2017mobilenets, sandler2018inverted, howard2019searching}, MNasNets~\cite{tan2018mnasnet} and EfficientNets~\cite{tan2019efficientnet} in Figure~\ref{figs:convergence_b}. We introduce a simple modification to this learning rate schedule, named \textit{exponentially decaying with constant ending}, which has a constant learning rate at the end of training when it reaches 5\% of the initial learning rate (Figure~\ref{figs:convergence_b}). It brings two benefits. First, with a slightly larger learning rate at the end, the small child models learn faster. Second, the constant learning rate at the end alleviates the overfitting of big child models as the weights oscillate.

\textbf{Regularization.} Empirically when comparing training/validation losses, we find big child models tend to overfit the training data whereas small child models tend to underfit. In previous work, Bender et al. \cite{bender2018understanding} apply the same weight decay to all child models regardless whether they are small or big. To prevent overfitting of larger networks, For EfficientNets, Tan et al. \cite{tan2019efficientnet} found it helpful to use larger dropout~\cite{srivastava2014dropout} rates for larger neural networks.
This becomes even more complicated in our context of training big single-stage models, due to the interplay among the small child models and big child models with shared parameters. Nevertheless, we introduce a simple rule that is surprisingly effective for this problem: \textit{regularize only the biggest (full) child model} (\ie, the only model that has direct access to the ground truth training labels since other child models are trained with inplace distillation only). We simply apply this rule to both weight decay and dropout, and empirically demonstrate its effectiveness in our experiments.

\textbf{Batch Norm Calibration.}
Batch norm statistics are not accumulated when training the single-stage model as they are ill-defined with varying architectures. After the training is completed, we re-calibrate the batch norm statistics (following Yu et al.~\cite{yu2019universally}) for each sampled child model for deployment without retraining or finetuning any network parameters.

\subsection{Coarse-to-fine Architecture Selection.}
After training a single-stage model, one needs to select the best architectures w.r.t.\ the resource budgets. Although obtaining the accuracy of a child model is cheap, the number of architecture candidates is extremely large (more than \(10^{12}\)). To address this issue, we propose a coarse-to-fine strategy where we first try to find a rough skeleton of promising network candidates in general, and then sample multiple fine-grained variations around each skeleton architecture of interest.

Specifically, in the coarse-grained phase, we define a limited input resolution set, depth set (global depth multipliers), channel set (global width multipliers) and kernel size set, and obtain benchmarks for all child models in this restricted space. This is followed by a fine-grained search phase, where we first pick the best network skeleton satisfying the given resource constraint found in the previous phase, and then randomly mutate its network-wise resolution, stage-wise depth, number of channels and kernel sizes to further discover better network architectures. Finally, we directly use the weights from the single-stage model for the induced child models without any retraining or finetuning. More details will be presented in the experiments.

\section{Experiments}
In this section, we first present the details of our search space, followed by our main results compared with the previous state-of-the-arts in terms of both accuracy and efficiency. Then we conduct an extensive ablative study to demonstrate the effectiveness of our proposed modifications. Finally, we show the intermediate results of our coarse-to-fine architecture selection.

\subsection{Search Space Definition}
\begin{table}[ht]
\small
\centering
\caption{MobileNetV2-based search space.}
\begin{tabular}{@{}c c c c c c@{}} \toprule
Stage & Operator & Resolution & \#Channels & \#Layers & Kernel Sizes\\
\midrule
 & Conv & \(192\times192\) - \(320\times320\) & 32 - 40 & 1 & 3\\
\cmidrule{1-6}
 1 & MBConv1 & \(96\times96\) - \(160\times160\) & 16 - 24 & 1 - 2 & 3\\
 \cmidrule{1-6}
 2 & MBConv6 & \(96\times96\) - \(160\times160\) & 24 - 32 & 2 - 3 & 3\\
 \cmidrule{1-6}
 3 & MBConv6 & \(48\times48\) - \(80\times80\) & 40 - 48 & 2 - 3 & 3, 5\\
 \cmidrule{1-6}
 4 & MBConv6 & \(24\times24\) - \(40\times40\) & 80 - 88 & 2 - 4 & 3, 5\\
 \cmidrule{1-6}
 5 & MBConv6 & \(12\times12\) - \(20\times20\) & 112 - 128 & 2 - 6 & 3, 5\\
 \cmidrule{1-6}
 6 & MBConv6 & \(12\times12\) - \(20\times20\) & 192 - 216 & 2 - 6 & 3, 5\\
 \cmidrule{1-6}
 7 & MBConv6 & \(6\times6\) - \(10\times10\) & 320 - 352 & 1 - 2 & 3, 5\\
 \cmidrule{1-6}
  &  Conv & \(6\times6\) - \(10\times10\) & 1280 - 1408 & 1 & 1 \\
\bottomrule
\end{tabular}
\label{tabs:search_space}
\end{table}

Following previous resource-aware NAS methods~\cite{tan2018mnasnet, tan2019efficientnet, cai2018proxylessnas, wu2019fbnet, howard2019searching}, our network architectures consist of a stack with inverted bottleneck residual blocks (MBConv)~\cite{sandler2018inverted}. We also insert a squeeze-and-excitation module~\cite{hu2018squeeze} in each block following EfficientNet~\cite{tan2019efficientnet} and MobileNetV3~\cite{howard2019searching}. The detailed search space is summarized in Table~\ref{tabs:search_space}. For the input resolution dimension, we sample from set \{192, 224, 288, 320\}. By training on different input resolutions, we find our trained single-stage model is able to generalize to unseen input resolutions during architecture search or deployment (\eg, 208, 240, 256, 272, 304, 336) after BN calibration. For the depth dimension, our network has seven stages (excluding the first and the last convolution layer). Each stage has multiple choices of the number of layers (\eg, stage 5 can pick any number of layers ranging from 2 to 6). Following slimmable networks~\cite{yu2018slimmable} that always keep lower-index channels in each layer, we always keep \textit{lower-index layers} in each network stage (and their weights). For weight sharing on the kernel size dimension in the inverted residual blocks, a \(3 \times 3\) depthwise kernel is defined to be the center of a \(5 \times 5\) depthwise kernel. Both kernel sizes and channel numbers can be adjusted layer-wise. The input resolution is network-wise and the number of layers is a stage-wise configuration in our search space.

\subsection{Main Results on ImageNet}
\begin{figure}[ht]
\centering
\includegraphics[width=\linewidth]{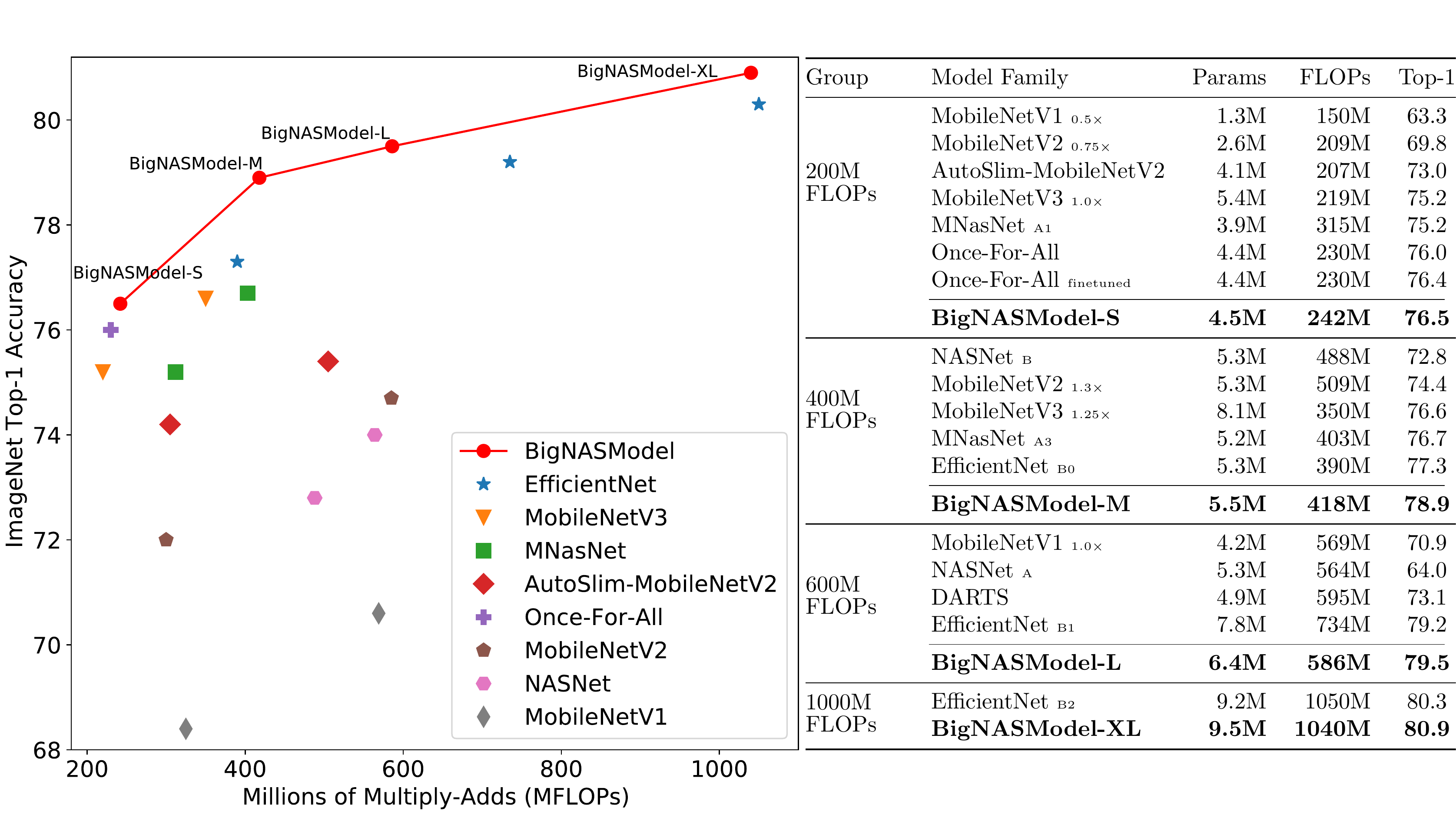}
\caption{Main results of BigNASModels on ImageNet. We note that MNasnet and MobileNetV3 are mainly optimized for on-device latency instead of FLOPs. The training hyper-parameters of these models are also different, and ours follow the baseline EfficientNets~\cite{tan2019efficientnet}.}
\label{figs:bignasnet}
\end{figure}

We train our big single-stage model on ImageNet~\cite{deng2009imagenet} using same settings following our strongest baseline EfficientNets~\cite{tan2019efficientnet}: RMSProp optimizer with decay 0.9 and momentum 0.9; batch normalization with post-calibration~\cite{yu2019universally}; weight decaying factor $10^{-5}$; initial learning rate 0.256 that decays by 0.97 every 2.4 epochs; swish activation~\cite{ramachandran2018searching} and AutoAugment policy~\cite{cubuk2019autoaugment}. We train our big single-stage model together with all techniques proposed in Section~\ref{secs:train_one_stage}. The learning rate is truncated to a constant value when it reaches 5\% of its initial value (\ie, 0.0128) until the training ends. We apply dropout only on training the full network with dropout ratio 0.2, applying weight decay only to the largest child model. To train the single-stage model, we adopt the sandwich sampling rules and inplace distillation proposed by~\cite{yu2019universally}. After the training, we use a simple coarse-to-fine architecture selection to find the best architecture under each interested resource budgets. We will show the details of coarse-to-fine architecture selection in Section~\ref{secs:grid_search}.

We show the performance benchmark of our model family, named BigNASModels, in Figure~\ref{figs:bignasnet}. On the left we show the visualization of FLOPs-Accuracy benchmarks compared with the previous arts including MobileNetV1~\cite{howard2017mobilenets}, NASNet~\cite{zoph2018learning}, MobileNetV2~\cite{sandler2018inverted}, AutoSlim-MobileNetV2~\cite{yu2019network}, MNasNet~\cite{tan2018mnasnet}, MobileNetV3~\cite{howard2019searching}, EfficientNet~\cite{tan2019efficientnet} and concurrent work Once-For-All~\cite{cai2019once}. We show the detailed benchmark results on the right table. For small-sized models, our BigNASModel-S achieves 76.5\% accuracy under only 240 MFLOPs, which is 1.3\% better than MobileNetV3 in terms of similar FLOPs, and 0.5\% better than ResNet-50~\cite{he2016deep} with \(17 \times\) fewer FLOPs. For medium-sized models, our BigNASModel-M achieves 1.6\% better accuracy than EfficientNet B0. For large-sized models where ImageNet classification accuracy saturates, our BigNASModel-L still has 0.6\% improvement compared with EfficientNet B2. Moreover, instead of individually training models of different sizes, our BigNASModel-S, BigNASModel-M and BigNASModel-L are sliced directly from one pretrained single-stage model, without retraining or finetuning.

\subsection{Ablation Study}

\begin{figure}[ht]
\centering
\includegraphics[width=0.48\linewidth]{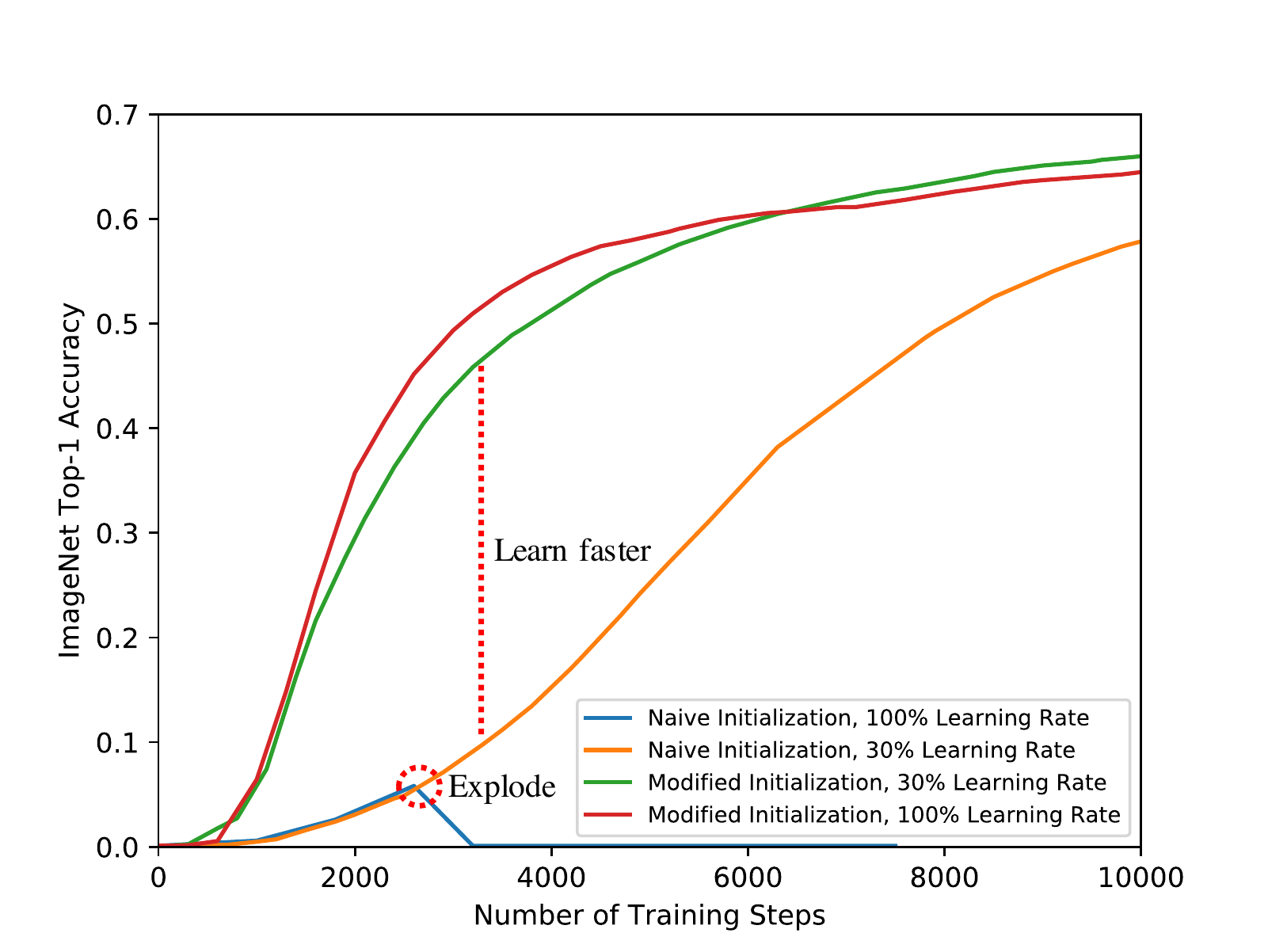}
\includegraphics[width=0.48\linewidth]{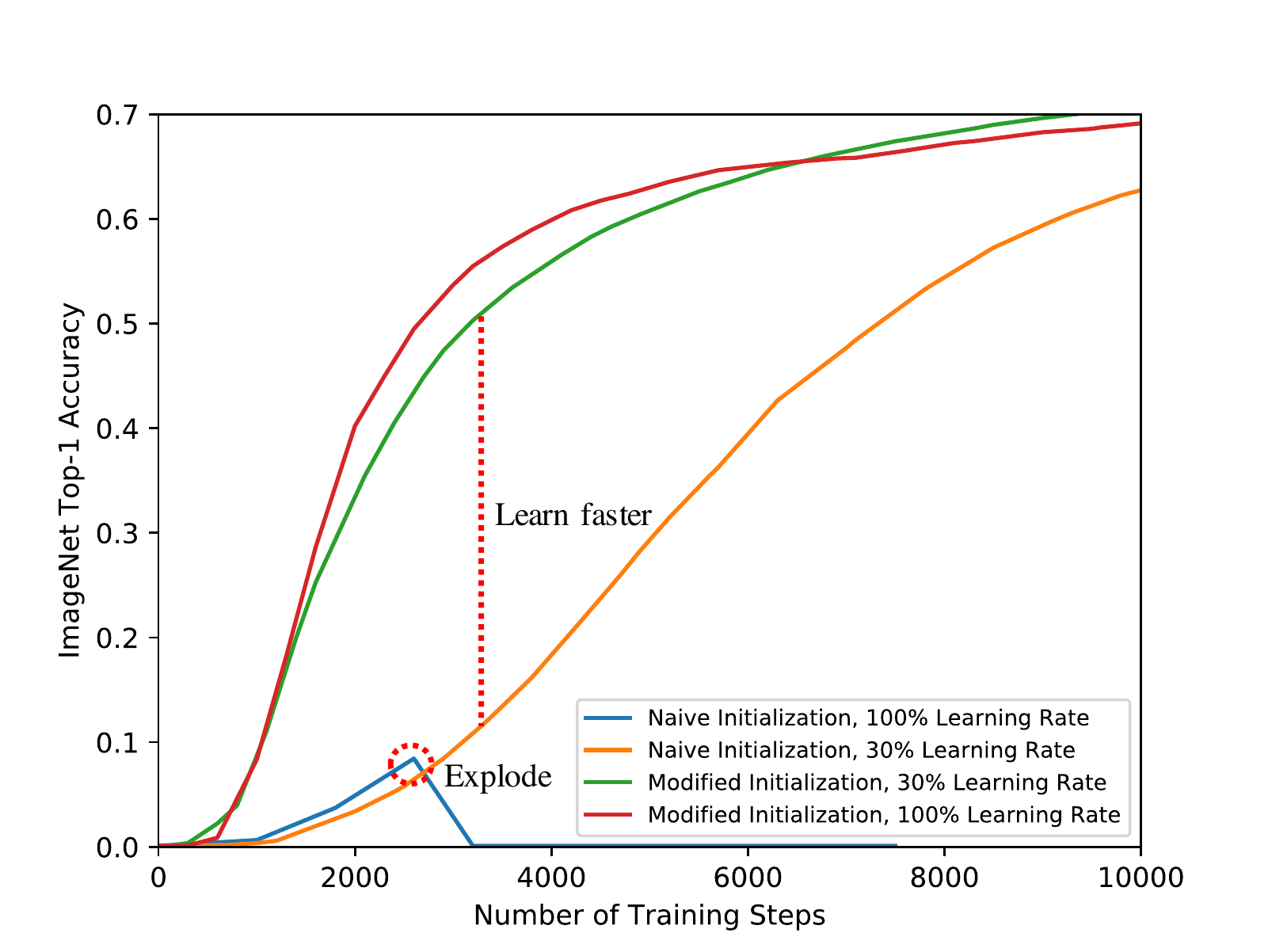}
\caption{\textbf{Focusing on the start of training.} Ablation study on different initialization methods. We show the validation accuracy of a small (left) and big (right) child model.}
\label{figs:initialization_start}
\end{figure}

\begin{figure}[ht]
\centering
\includegraphics[width=0.48\linewidth]{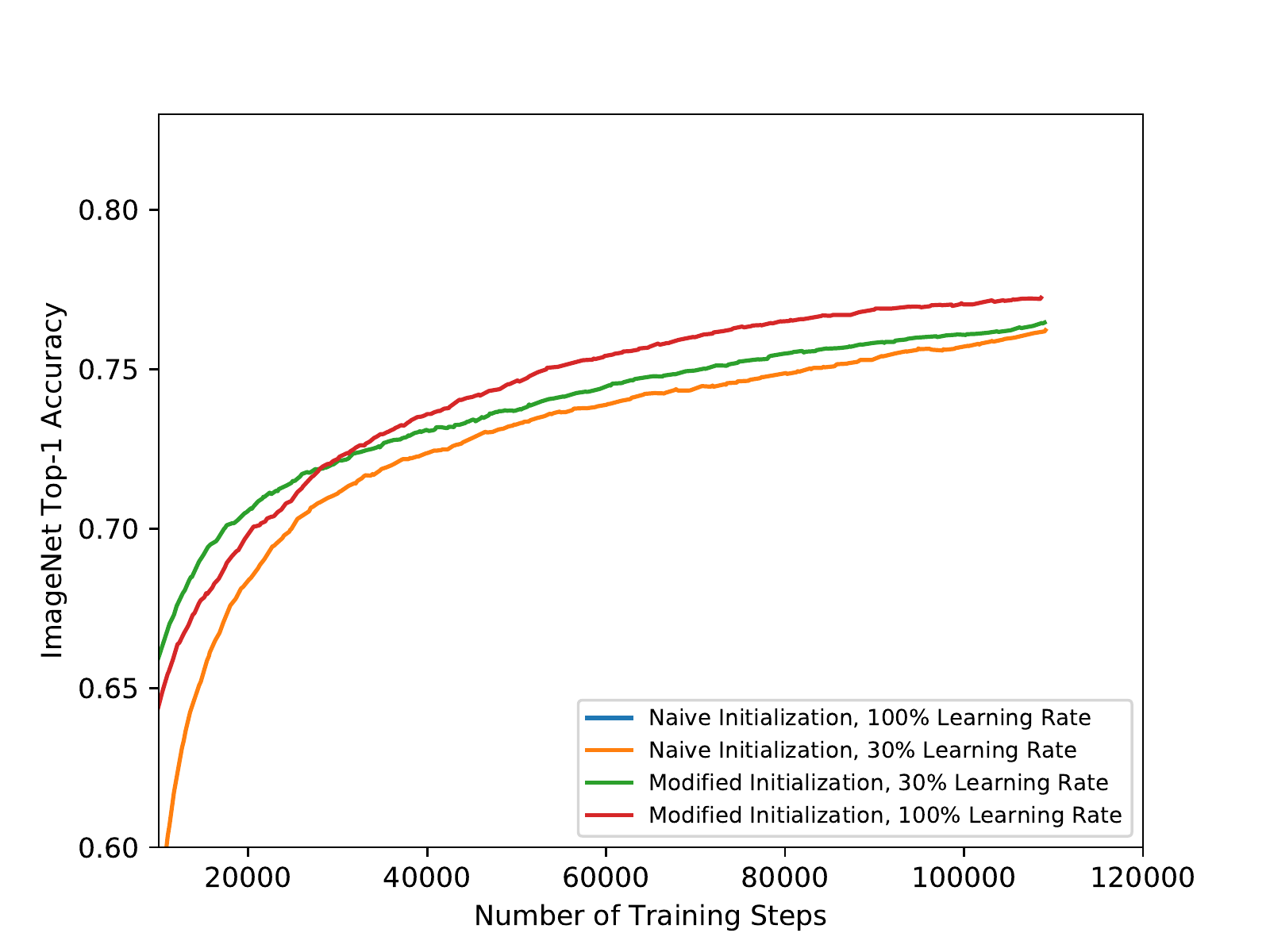}
\includegraphics[width=0.48\linewidth]{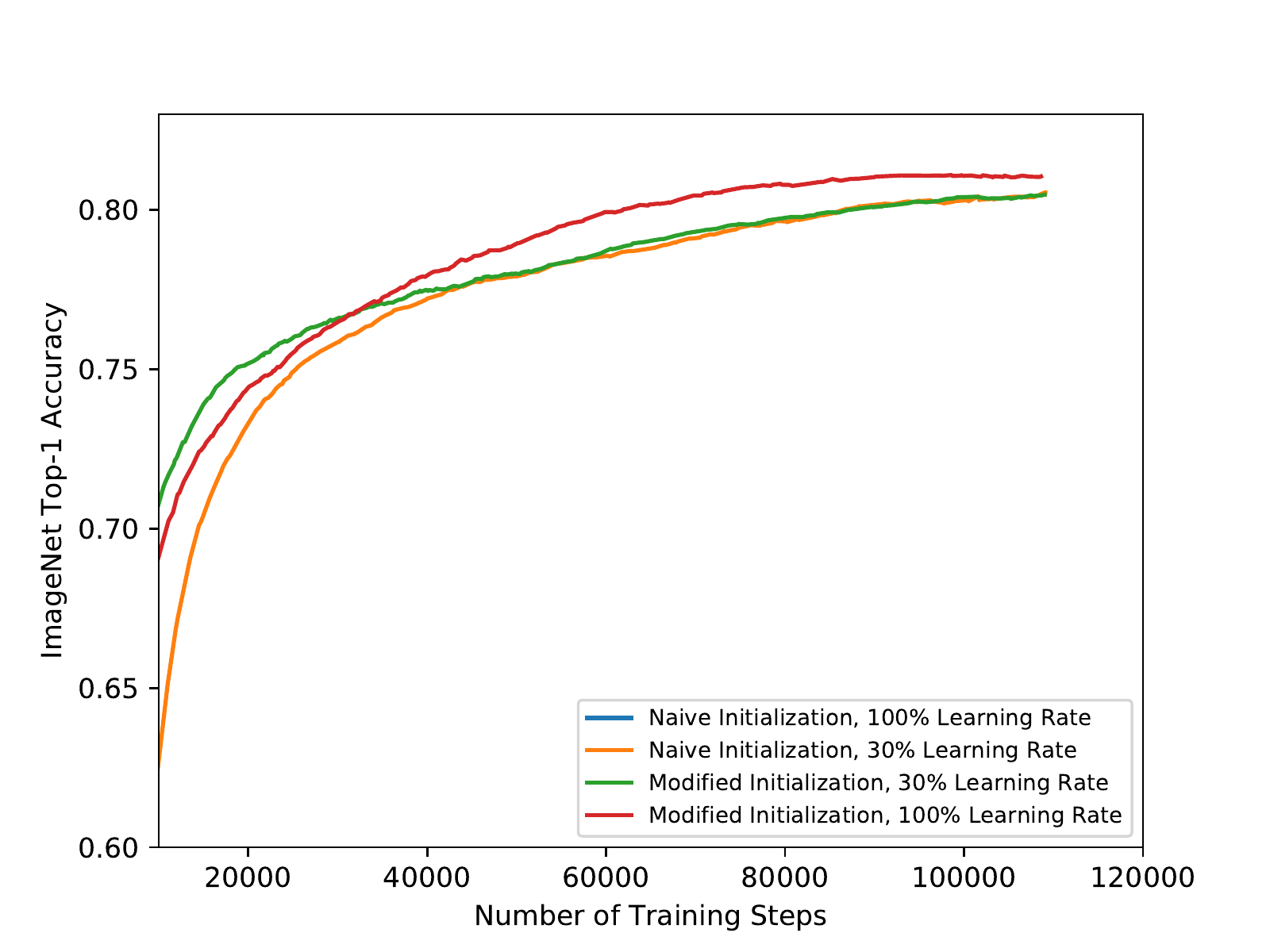}
\caption{\textbf{Focusing on the end of training.} Ablation study on different initialization methods. We show the validation accuracy of a small (left) and big (right) child model.}
\label{figs:initialization_end}
\end{figure}

\textbf{Ablation Study on Initialization.}
Previous weight initialization methods~\cite{he2015delving} are deduced from fixed neural networks, where the numbers of input units is constant. However, in a single-stage model, the number of input units varies across the different child models. In this part, we start with training a single-stage model using He Initialization~\cite{he2015delving} designed for fixed neural networks. As shown in Figure~\ref{figs:initialization_start}, the accuracy of both small (left) and big (right) child models drops to zero after a few thousand training steps during the learning rate warming-up~\cite{goyal2017accurate}. The single-stage model is able to converge when we reduce the learning rate to the 30\% of its original value. If the initialization is modified according to Section~\ref{secs:train_one_stage}, the model learns much faster at the beginning of the training (shown in Figure~\ref{figs:initialization_start}), and has better performance at the end of the training (shown in Figure~\ref{figs:initialization_end}). Moreover, we are also able to train the single-stage model with the original learning rate hyper-parameter, which leads to much better performance for both small (Figure~\ref{figs:initialization_end}, left) and big (Figure~\ref{figs:initialization_end}, right) child models.

\begin{figure}[hb]
\centering
\includegraphics[width=\linewidth]{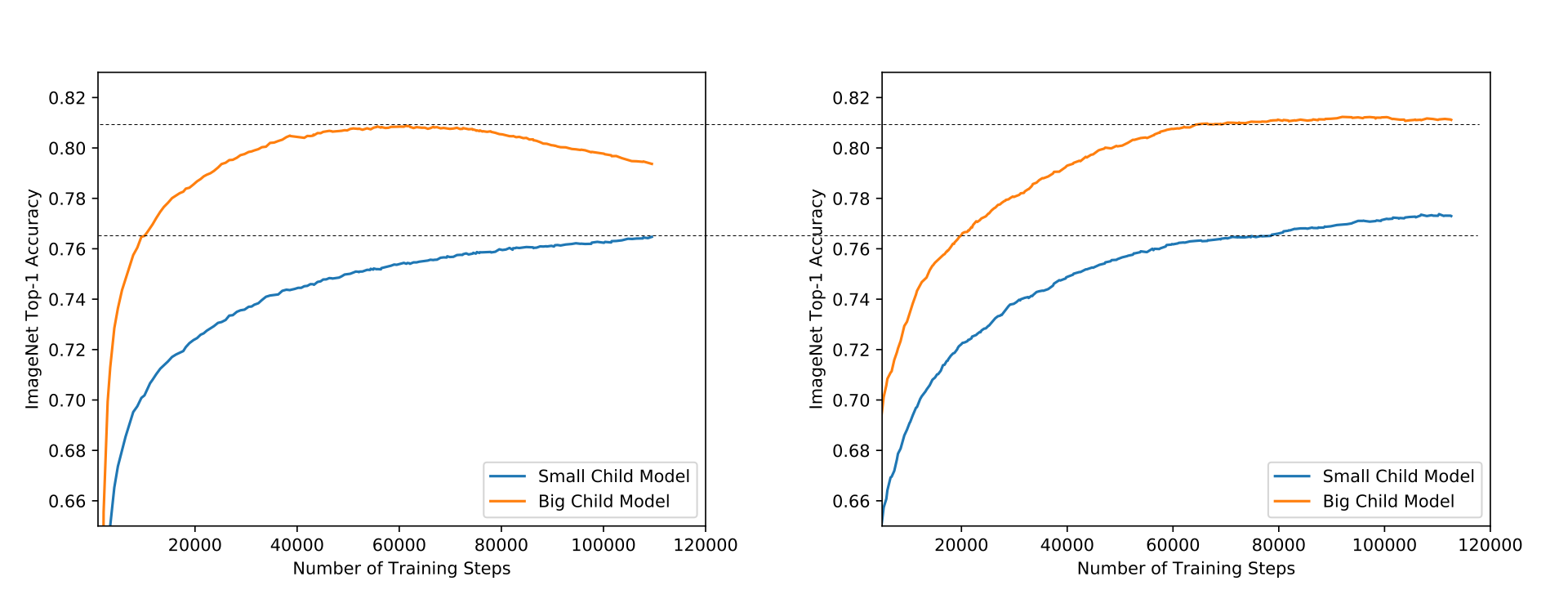}
\caption{The validation accuracy curves during the training process for both small and big child models before (left) and after (right) our modifications.}
\label{figs:ablation_convergence}
\end{figure}

\textbf{Ablation Study on Convergence Behavior.}
During the training of a single-stage model, the big child models converge faster and then overfit, while small child models converge slower and need more training. In this part, we show the performance after addressing this issue in Figure~\ref{figs:ablation_convergence}. We apply the proposed learning rate schedule \textit{exponentially decaying with constant ending} on the right. The detailed learning rate schedules are shown in Figure~\ref{figs:convergence_b}. We also tried many other learning rate schedules with an exhaustive hyper-parameter sweep, including linearly decaying~\cite{ma2018shufflenet, yu2019universally} and cosine decaying~\cite{loshchilov2016sgdr, he2019bag}. But the performances are all worse than exponentially decaying.

\begin{figure}[ht]
\centering
\includegraphics[width=0.48\linewidth]{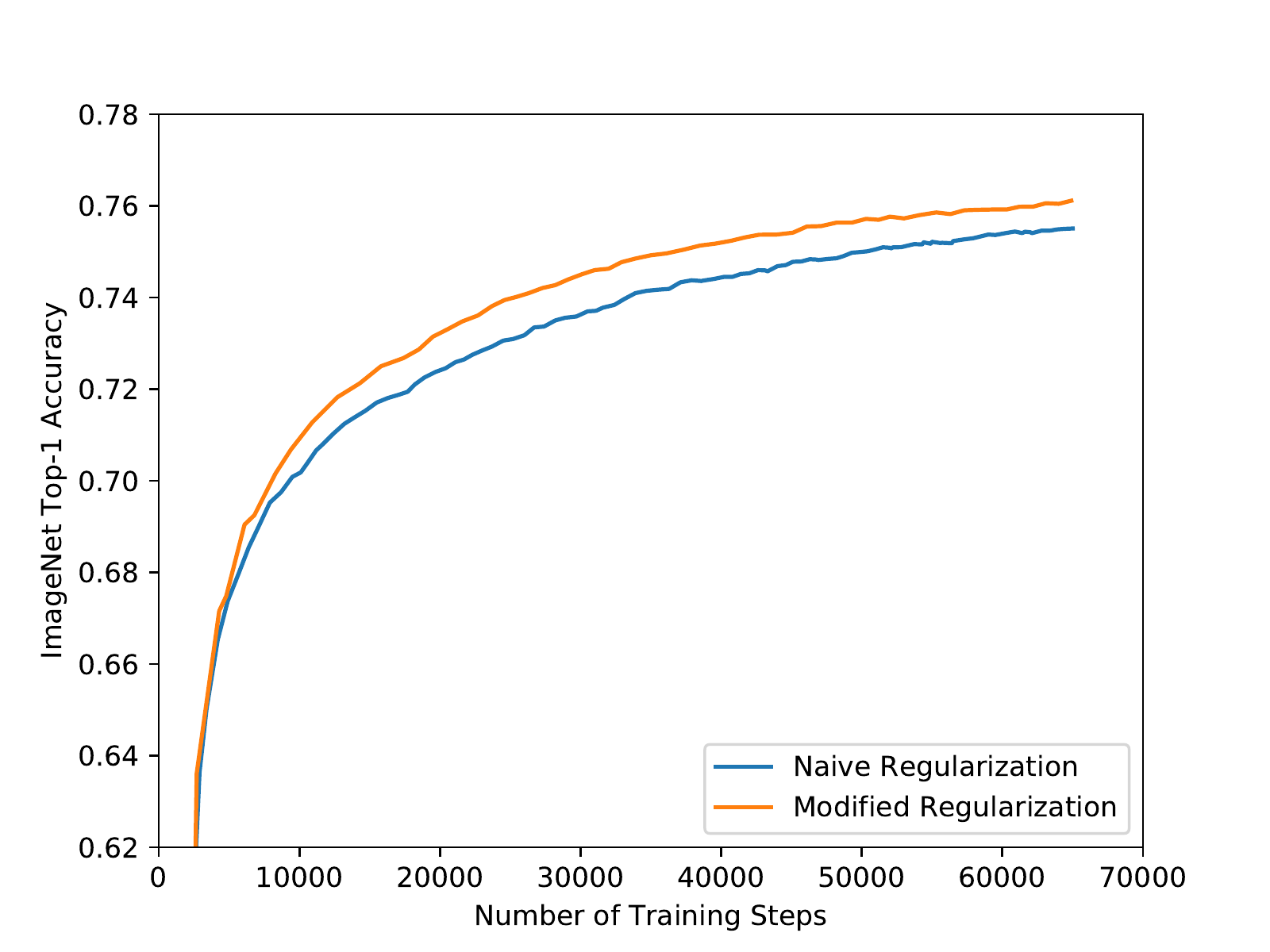}
\includegraphics[width=0.48\linewidth]{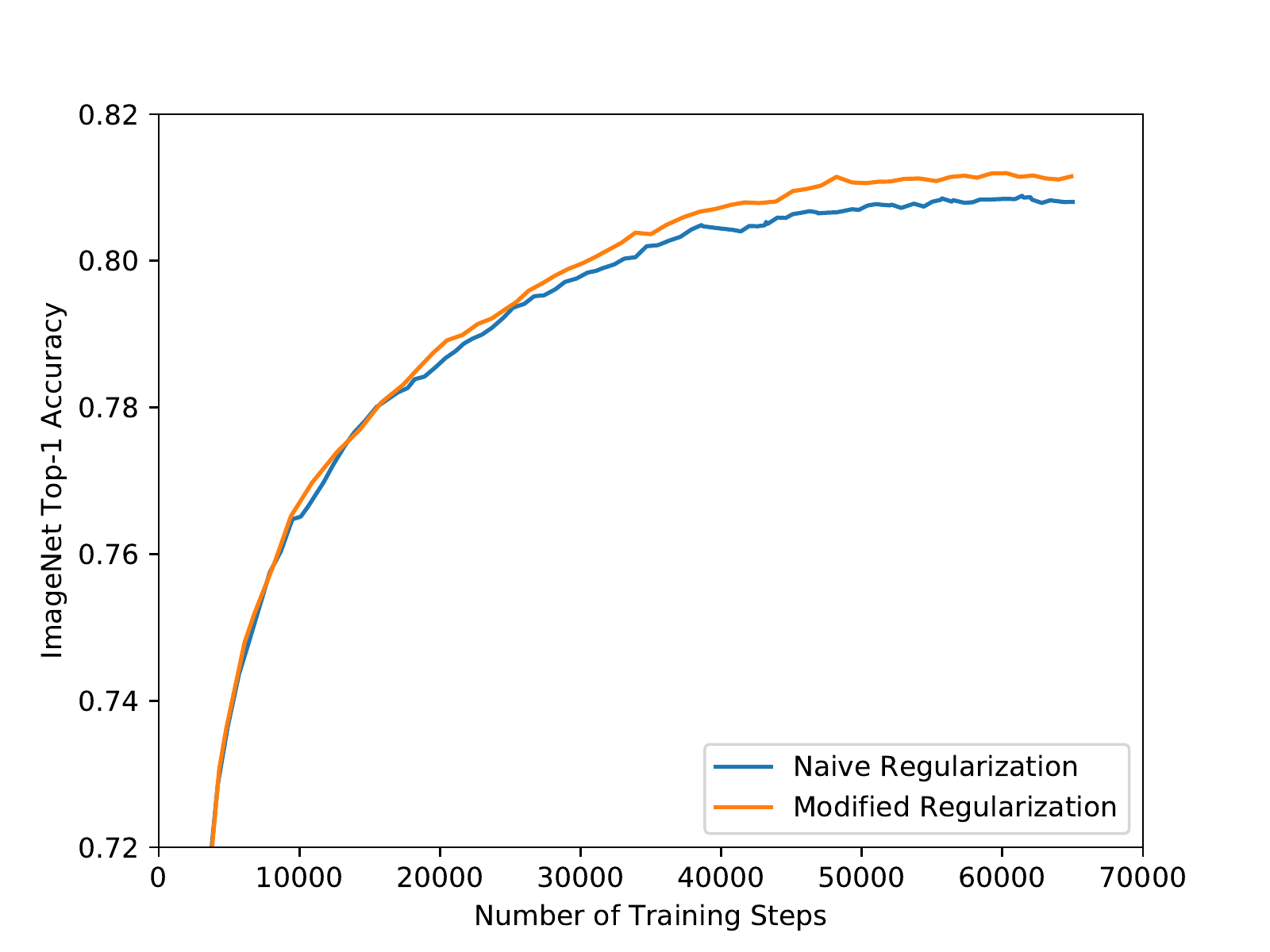}
\caption{The validation accuracy of a small (left) and big (right) child model using different regularization rules.}
\label{figs:regularization}
\end{figure}

\textbf{Ablation Study on Regularization.}
Big child models are prone to overfitting on the training data whereas small child models are prone to underfitting. In this part, we compare the effects of the regularization between two rules: (1) applying regularization on all child models~\cite{bender2018understanding}, and (2) applying regularization only on the full network. Here the regularization techniques we consider are weight decay with factor $10^{-5}$ and dropout with ratio \(0.2\) (the same hyper-parameters used in training previous state-of-the-art mobile networks). Figure~\ref{figs:regularization} shows the performance of both small (left) and big (right) child models using different regularization rules. On the left, the performance of small child models is improved by a large margin (\(+0.5\) top-1 accuracy) as they have less regularization and more capacity to fit the training data. Meanwhile on the right, we found the performance of the big child model is also improved slightly (\(+0.2\)).

\subsection{Coarse-to-fine Architecture Selection}\label{secs:grid_search}

\begin{figure}[ht]
\centering
\includegraphics[width=\linewidth]{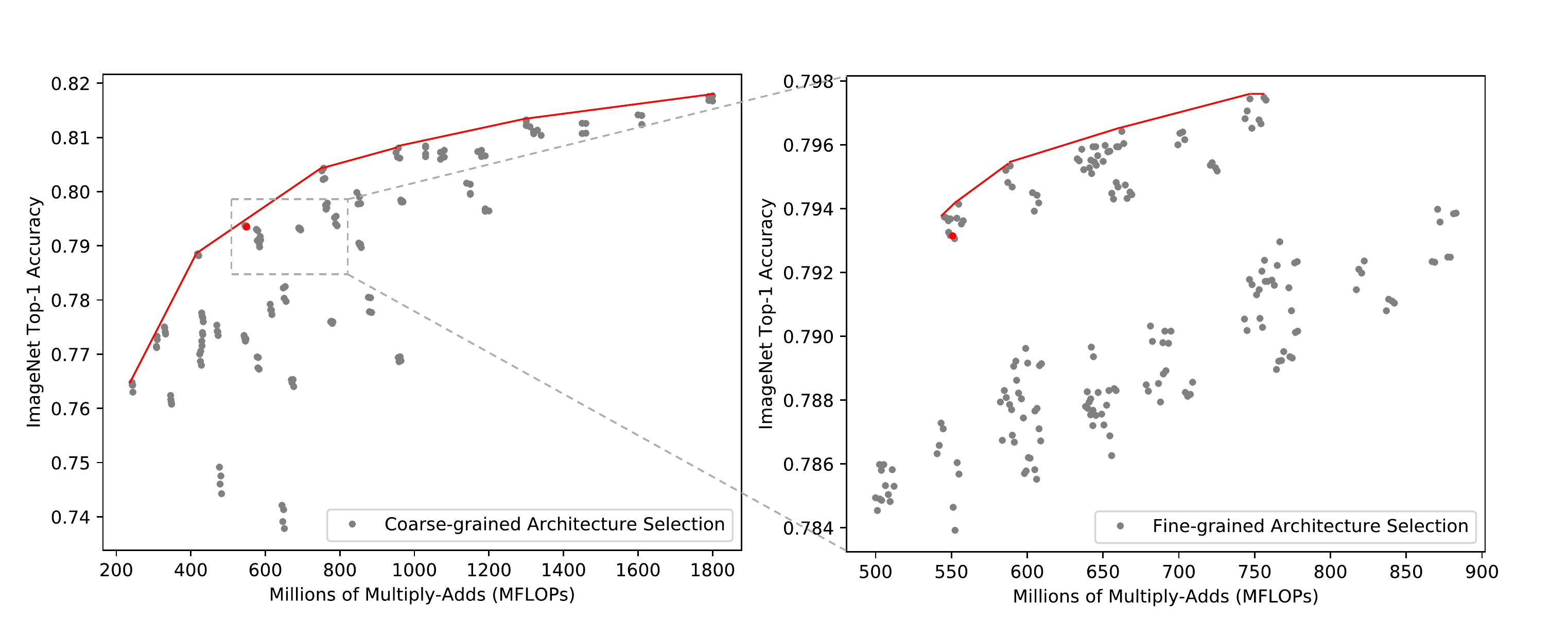}
\caption{Results of coarse-to-fine architecture selection. The red dot in coarse-grained architecture selection is picked and mutated for fine-grained  selection.
}
\label{figs:grid_search}
\end{figure}

After the training of a single-stage model, we use coarse-to-fine architecture selection to find the best architectures under different resource budgets. During the search, the evaluation metrics can be flexible including predictive accuracy, FLOPs, memory footprint, latency on various different devices, and many others. It is noteworthy that we pick the best architectures according to the predictive accuracy on training set, because we used all training data for obtaining our single-stage model (no retraining from scratch), and the validation set of ImageNet~\cite{deng2009imagenet} is being used as ``test set'' in the community. In this part, we first show an illustration of our coarse-to-fine architecture selection with the trained big single-stage model in Figure~\ref{figs:grid_search}. The search results are based on FLOPs-Accuracy benchmarks (as FLOPs are more reproducible and independent of the software version, hardware version, runtime environments and many other factors).

During the coarse-to-fine architecture selection, we first find rough skeletons of good candidate networks. Specifically, in the coarse selection phase, we pre-define five input resolutions (network-wise, {\tiny\{192, 224, 256, 288, 320\}}), four depth configurations (stage-wise via global depth multipliers~\cite{tan2019efficientnet}), two channel configurations (stage-wise via global width multipliers~\cite{howard2017mobilenets}) and four kernel size configurations (stage-wise), and obtain all of their benchmarks (shown in Figure~\ref{figs:grid_search} on the left). Then under our interested latency budget, we perform a fine-grained grid search by varying its configurations (shown in Figure~\ref{figs:grid_search} on the right). For example, under FLOPs near 600M we first pick the skeleton of the red dot shown in Figure~\ref{figs:grid_search}. We then perform additional fine-grained architecture selection by randomly varying the input resolutions, depths, channels and kernel sizes slightly. We note that the coarse-to-fine architecture selection is flexible and not very exhaustive in our experiments, yet it already discovered fairly good architectures as shown in Figure~\ref{figs:grid_search} on the right. For the FLOPs near 650M, we finally select the child model with input resolution 256, depth configuration {\tiny\{1:2:2:2:4:4:1\}}, channel configuration {\tiny\{32:16:24:48:88:128:216:352:1408\}} and kernel size configuration {\tiny\{3:3:5:3:5:5:3\}}. After training of the single-stage model, the post-search step is highly parallelizable and independent of training.

\section{Analysis of BigNASModel}
\noindent\hspace{\parindent} \textbf{Finetuning child models sampled from BigNASModel.}
In previous sections we have reported the accuracies of child models from a single trained BigNASModel without finetuning, what if we do finetune it? To understand whether the trained BigNASModel has reached relatively optimal accuracies, we conduct experiments to finetune these child models (\ie, BigNASModel-S, BigNASModel-M, BigNASModel-L, BigNASModel-XL) for additional 25 epochs under different constant learning rates separately. Table~\ref{tabs:finetune} shows that finetuning in our setting no longer improves accuracy significantly.
\begin{table}[ht]
\centering
\caption{Analysis on Child Models sampled from BigNASModel. We compare the ImageNet validation performance of (1) child model directly sampled from BigNASModel without finetuning (\textbf{w/o} Finetuning), (2) child model finetuned with various constant learning rate (\textbf{w/} Finetuning at different lr). {\color{blue} Blue} subscript indicates the performance improvement while {\color{red} Red} subscript indicates degradation.}
\begin{tabular}{@{}l c | c | c | c | c @{}} \toprule
Child Model & & \textbf{w/o} Finetuning  & \textbf{w/} Fintuning & \textbf{w/} Fintuning  & \textbf{w/} Fintuning\\
& & & {\small lr = 0.01} & {\small lr = 0.001} & {\small lr = 0.0001}\\
\midrule
BigNASModel-S && 76.5 & 74.6 \textsubscript{\color{red} (-1.9)} & 76.4 \textsubscript{\color{red} (-0.1)} & 76.5 \textsubscript{(0.0)}\\
BigNASModel-M && 78.9 & 76.7 \textsubscript{\color{red} (-2.2)} & 78.8 \textsubscript{\color{red} (-0.1)} & 78.8 \textsubscript{\color{red} (-0.1)}\\
BigNASModel-L && 79.5 & 77.9 \textsubscript{\color{red} (-1.6)} & 79.6 \textsubscript{\color{blue} (+0.1)} & 79.7 \textsubscript{\color{blue} (+0.2)}\\
BigNASModel-XL && 80.9 & 79.0 \textsubscript{\color{red} (-1.9)} & 80.6 \textsubscript{\color{red} (-0.3)} & 80.8 \textsubscript{\color{red} (-0.1)}\\
\bottomrule
\end{tabular}
\label{tabs:finetune}
\end{table}
\begin{table}[ht]
\centering
\vspace{-5mm}
\caption{Analysis on training child architectures from scratch. We compare the ImageNet validation accuracy of (1) child model directly sampled from BigNASModel without finetuning (\textbf{w/o} Finetuning), (2) child architectures trained from scratch without distillation (FromScratch \textbf{w/o} distill), and (3) child architectures trained from scratch with two distillation methods A~\cite{hinton2015distilling} and B~\cite{yu2019universally} (FromScratch \textbf{w/} distill (A)/(B)).}
\begin{tabular}{@{}l c | c | c | c | c @{}} \toprule
Child Architecture & & \textbf{w/o} Finetuning & FromScratch & FromScratch & FromScratch \\
& & & \textbf{w/o} distill & \textbf{w/} distill (A) & \textbf{w/} distill (B)\\
\midrule
BigNASModel-S && 76.5 & 75.3 \textsubscript{\color{red} (-1.2)} & 75.3 \textsubscript{\color{red} (-1.2)} & 76.3 \textsubscript{\color{red} (-0.2)} \\
BigNASModel-M && 78.9 & 77.4 \textsubscript{\color{red} (-1.5)} & 77.4 \textsubscript{\color{red} (-1.5)} & 78.6 \textsubscript{\color{red} (-0.3)}\\
BigNASModel-L && 79.5 & 78.2 \textsubscript{\color{red} (-1.3)} & 77.9 \textsubscript{\color{red} (-1.5)} & 79.2 \textsubscript{\color{red} (-0.3)}\\
BigNASModel-XL && 80.9 & 79.3 \textsubscript{\color{red} (-1.6)} & 79.0 \textsubscript{\color{red} (-1.9)} & 80.4 \textsubscript{\color{red} (-0.5)}\\
\bottomrule
\end{tabular}
\label{tabs:fromscratch}
\vspace{-5mm}
\end{table}

\noindent\hspace{\parindent}
\textbf{Training the architectures of child from scratch.} We further study the performance when these selected child models are trained from scratch with or without distillation. We implement two distillation variants. The first distillation, referred as Distill (A), is a simple distillation~\cite{hinton2015distilling} without temperature. The teacher network is trained with dropout and label smoothing following our training pipeline. The student network is trained with distillation loss. The second distillation method, referred as Distill (B), is inplace distillation~\cite{yu2019universally} where we jointly train a teacher and student network from scratch with weight sharing. The student network is trained with the soft-predictions of the teacher network only. The Distill (B) is most similar to the distillation used in training BigNASModel. We note that although it is commonly believed that distillation can improve regularization, we found that the simple Distill (A) method does not help in EfficientNet-based architectures. Table~\ref{tabs:fromscratch} shows that the accuracies of child models slightly benefit from jointly training a weight-sharing single-stage model, which is consistent to the observations in previous work~\cite{yu2019universally}.

\section{Conclusion}
We presented a novel paradigm for neural architecture search by training a single-stage model, from which high-quality child models of different sizes can be induced for instant deployment without retraining or finetuning. With several proposed techniques, we obtain a family of BigNASModels as slices in a big pre-trained single-stage model. These slices simultaneously surpass all state-of-the-art ImageNet classification models ranging from 200 MFLOPs to 1 GFLOPs. We hope our work can serve to further simplify and scale up neural architecture search.

\clearpage

\bibliographystyle{splncs04}
\bibliography{egbib}

\title{BigNAS: Scaling Up Neural Architecture Search with Big Single-Stage Models\\
\vspace{2mm}
(Supplementary Materials)}

\titlerunning{BigNAS: Neural Architecture Search with Big Single-Stage Models}

\author{Jiahui Yu\inst{1,2} \and
Pengchong Jin\inst{1} \and
Hanxiao Liu\inst{1} \and
Gabriel Bender\inst{1} \and\\
Pieter-Jan Kindermans\inst{1} \and
Mingxing Tan\inst{1} \and
Thomas Huang\inst{2} \and
Xiaodan Song\inst{1} \and
Ruoming Pang\inst{1} \and
Quoc Le\inst{1}
}

\authorrunning{J. Yu et al.}

\institute{Google Brain \and University of Illinois at Urbana-Champaign \\\email{jiahuiyu@google.com}}

\maketitle
\appendix

\section{Architectures of BigNASModel}
\begin{figure}[ht]
\centering
\includegraphics[width=0.98\linewidth]{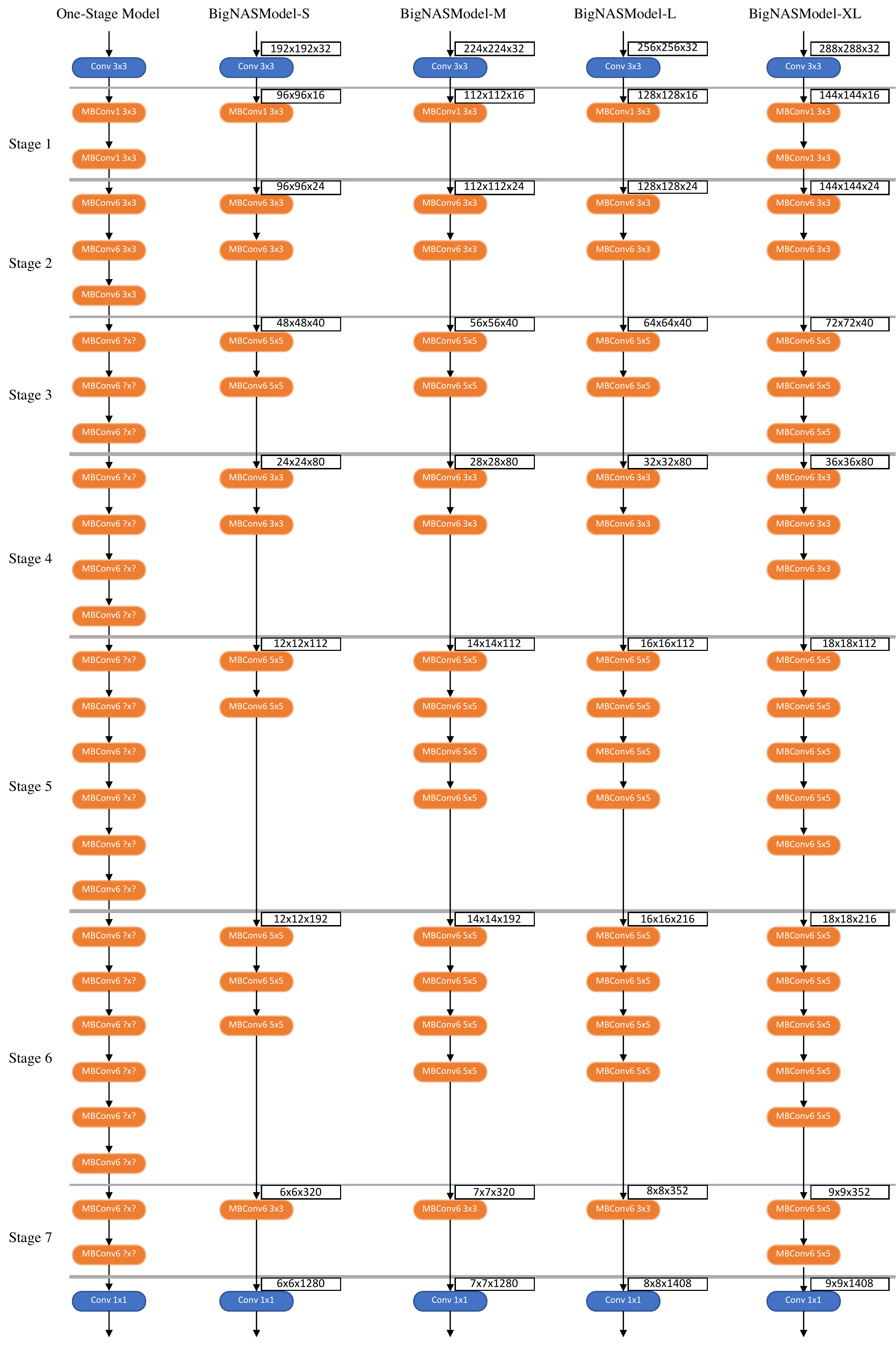}
\caption{Architecture visualization of the single-stage model and child models BigNASModel-S, BigNASModel-M, BigNASModel-L, BigNASModel-XL. All child models are directly sliced from the single-stage model without retraining or finetuning.}
\label{figs:vis_architecture}
\end{figure}

We show the architecture visualization of the single-stage model and child models BigNASModel-S, BigNASModel-M, BigNASModel-L, BigNASModel-XL in Figure~\ref{figs:vis_architecture}. The child models are directly sliced from the single-stage model without retraining or finetuning. Compared with the compound model scaling heuristic~\cite{tan2019efficientnet}, our child models have distinct architectures across all dimensions. For example, comparing BigNASModel-XL with EfficientNet-B2, the EfficientNet-B2 has input resolution 260,  channels {\tiny\{40:24:32:40:88:128:216:352:1408\}}, kernel sizes {\tiny\{3:3:5:3:5:5:3\}} and stage layers {\tiny\{2:3:3:4:4:5:2\}}. Our BigNASModel-XL achieves 80.9\% top-1 accuracy under 1040 MFLOPs, while EfficientNet-B2 achieves 80.3\% top-1 accuracy under 1050 MFLOPs.

\section{Learning Rate Schedule: Exponentially Decaying with Constant Ending}
To address the distinct convergence behaviors among small and big child models, we proposed to train with a constant ending learning rate where we pick the 5\% of the initial learning rate as the minimum. We note that 5\% was meant to be a small constant value and was not specifically tuned. We conducted additional experiments in this section and verified that the results are insensitive \wrt this hyper-parameter. For example, we trained a weight-shared model based on small and big EfficientNet with different minimum learning rate values: 3\%, 5\%, 8\%, 10\% and the average performance is similar as shown in Table~\ref{tabs:learning_rate}.

\begin{table}[ht]
\centering
\caption{Exponentially Decaying with Constant Ending learning rate schedule. We trained a weight-shared model based on small and big EfficientNet with different minimum learning rate values: 3\%, 5\%, 8\%, 10\% and the average Top-1 accuracy is similar.}
\begin{tabular}{@{}l c  c  c  c @{}} \toprule
\% of initial LR & & Smallest Model  & Biggest Model & Average \\
\midrule
3\% && 76.5 & 80.8 & 78.7\\
5\% && 76.4 & 81.1 & 78.8\\
8\% && 76.3 & 81.3 & 78.8\\
10\% && 76.3 & 81.3 & 78.8\\
\bottomrule
\end{tabular}
\label{tabs:learning_rate}
\end{table}

\section{Implementation Details}

We implement all training and coarse-to-fine architecture selection algorithms on TensorFlow framework~\cite{tensorflow2015-whitepaper}. All of our experiments are conducted on \(8 \times 8\) TPUv3 pods. For ImageNet experiments, we use a total batch size 4096. Our single-stage model has sizes from 200 to 2000 MFLOPs, from which we search architectures from 200 to 1000 MFLOPs. To train a single-stage model, it roughly takes 36 hours.

Training on TPUs requires defining a static computational graph, where the shapes of all tensors in that graph should be fixed. Thus, during the training we are not able to dynamically slice the weights, select computational paths or sample many input resolutions. To this end, here we provide the details of our implementation for training single-stage models on TPUs. On the dimensions of kernel sizes, channels, and depths, we use the masking strategy to simulate the weight slicing or path selection during the training (\ie, we mask out the rest of the channels, kernel paddings, or the entire output of a residual block). On the dimension of input resolutions, in each training iteration, our data pipeline provides same images with four fixed resolutions (\{192, 224, 288, 320\}) which are paired with the model sizes. The smallest child model is always trained on the lowest resolution, while the biggest child model is always trained on the highest resolution. For all other resolutions the models are randomly varied on kernel sizes, channels, and depths. By this implementation, our trained single-stage model is able to provide high-quality child models across all these dimensions. For inference, we directly declare a child model architecture and load the sliced weights from the single-stage model. To slice the weights, we always use lower-index channels in each layer, lower-index layers in each stage, and the center \(3 \times 3\) depthwise kernel from a \(5 \times 5\) depthwise kernel.

For the data prefetching pipeline, we need multiple image input resolutions during the training. We first prefetch a batch of training patches with a fixed resolution (on ImageNet we use 224) with data augmentations, and then resize them with bicubic interpolation to our target input resolutions (\eg, 192, 224, 288, 320). We note that during inference, the single-crop testing accuracy is reported. Importantly, for testing data prefetching pipeline, we also prefetch a 224 center crop first and then resize to the target resolution to avoid the inconsistency.

During the training, we use cross-replica (synchronized) batch normalization following EfficientNets~\cite{tan2019efficientnet}. To enable this, we also have to use stateless random sampling function~\footnote{\url{https://www.tensorflow.org/api_docs/python/tf/random/stateless_uniform}} since naive random sampling function~\footnote{\url{https://www.tensorflow.org/api_docs/python/tf/random/uniform}} leads to different sampled values across different TPU cores. The input seed of stateless random sampling functions is the global training step plus current layer index so that the trained single-stage model can provide child models with different layer-wise/stage-wise configurations.

\end{document}